# Threat Detection in Social Media Networks Using Machine Learning Based Network Analysis

Aditi Sanjay Agrawal[1]
[1]Researcher, MSCS UC Davis CA, United States

Corresponding Author *   Aditi.agrawal2112@gmail.com

**Abstract**
The accelerated development of social media websites has posed intricate security issues in cyberspace, where these sites have increasingly become victims of criminal activities including attempts to intrude into them, abnormal traffic patterns, and organized attacks. The conventional rule-based security systems are not always scalable and dynamic to meet such a threat. This paper introduces a threat detection framework based on machine learning that can be used to classify malicious behavior in the social media network environment based on the nature of network traffic. Exploiting a rich network traffic dataset, the massive preprocessing and exploratory data analysis is conducted to overcome the problem of data imbalance, feature inconsistency, and noise. A model of artificial neural network (ANN) is then created to acquire intricate, non-linear tendencies of malicious actions. The proposed model is tested on conventional performance metrics, such as accuracy, accuracy, recall, F1-score, and ROC-AUC, and shows good detection and high levels of strength. The findings suggest that neural network-based solutions have the potential to be used effectively to identify the latent threat dynamics within the context of a large-scale social media network and that they can be employed to complement the existing intrusion detection system and better to conduct proactive cybersecurity operations.

1. Introduction

The widespread adoption of social media platforms has transformed the way individuals and organizations communicate, share information, and conduct digital interactions. However, this rapid growth has also made social media networks attractive targets for cyber threats, including unauthorized intrusions, malicious traffic propagation, and coordinated attack behaviors. Due to the scale, heterogeneity, and dynamic nature of social media environments, ensuring robust security has become increasingly challenging.

Traditional intrusion detection systems (IDS) largely rely on rule-based or signature-driven techniques, which are effective only against known attack patterns. These approaches struggle to adapt to evolving threat landscapes, particularly in large-scale social media networks where attack behaviors continuously change in form and intensity. As a result, there is a growing need for intelligent, adaptive security mechanisms capable of identifying complex and previously unseen threats.

Machine learning–based intrusion detection has emerged as a promising alternative, offering the ability to learn patterns directly from network traffic data. By leveraging statistical and behavioral features, machine learning models can distinguish between normal and malicious activities with greater flexibility than conventional methods. Among these approaches, artificial neural networks (ANNs) are particularly effective due to their capacity to model non-linear relationships and capture intricate interactions among network features.

In this study, we propose a machine learning–driven threat detection framework designed for social media network environments. The framework utilizes detailed network traffic characteristics to identify malicious behavior through supervised learning. Extensive exploratory data analysis and preprocessing are conducted to address challenges such as class imbalance, feature inconsistency, and data noise. An ANN-based classification model is then developed to learn threat patterns from the processed data.

1. The primary contributions of this work are as follows:
2. A comprehensive preprocessing and exploratory analysis pipeline tailored for large-scale network traffic data.
3. The design and implementation of an ANN-based threat detection model capable of learning complex attack patterns.
4. An empirical evaluation of the proposed framework using standard performance metrics to demonstrate its effectiveness and robustness.

## 2. Literature Survey

Machine learning and deep learning techniques have been extensively applied to pattern recognition and anomaly detection problems across diverse domains. Convolutional and probabilistic neural networks have demonstrated strong performance in extracting discriminative features from complex and noisy data, as evidenced by applications in agricultural disease detection and real-time biometric systems. Such studies highlight the ability of neural models to learn non-linear representations critical for accurate classification in real-world environments. In online communication systems, smart systems have been worked out to streamline and improve user interactions. The history before the current development of implementing AI-based applications into team-based work platforms like Slack indicates the increased dependence on machine learning models in social and communication networks, with a strong need to develop effective and safe design systems.

In more recent times, more direct research has been directed at defining social threats on the online platform. Threat detection on social media sites like X (previously Twitter) from sentiments has been investigated by utilizing supervised machine learning models including support vector machine and long short-term memory networks [1,2]. Although these methods are efficient to capture semantic and emotional signs of any malicious intent, they mainly work based on content level analysis and are prone to the language diversity and the ambiguity of a situation.

Network-based intrusion detection, on the other hand, concentrates on textual content without paying attention to behavioral characteristics of traffic flows. Despite the potential of neural network-based intrusion detection systems as learning complex traffic patterns, there is a lack of research on the applicability of neural network-based intrusion detection system in the scale of a social network. This research is based on the previous works in the field of deep learning by introducing an ANN-based threat detection model that processes network traffic characteristics that supplement content-based methods and adds an extra layer of security to social media websites.

The dataset we use in this research is the UNSW-NB15 dataset, a recent, state-of-the-art benchmark dataset that is used to test network intrusion detection systems. The IXIA Perfect Storm tool was used to create the dataset to model realistic network traffic that was generated with both benign traffic and a wide variety of modern cyber-attacks. This is because unlike older datasets, UNSW-NB15 captures current network dynamics and attack patterns and thus is applicable in research that deals with large-scale and dynamic network systems, i.e., social media platforms.

| | Attack category | Attack subcategory | Number of events |
|---|---|---|---|
| 0 | normal | NaN | 2218761 |
| 1 | Fuzzers | FTP | 558 |
| 2 | Fuzzers | HTTP | 1497 |
| 3 | Fuzzers | RIP | 3550 |
| 4 | Fuzzers | SMB | 5245 |

Fig.1 Attack Data

The dataset consists of multiple subsets that are merged to form a comprehensive corpus of network flow records. After data concatenation, duplication, and cleaning, the final dataset contains over two million records, each described by 49 features. These features include a combination of numerical and categorical attributes capturing flow-level, packet-level, protocol-level, and temporal characteristics of network traffic. Examples include flow duration, packet counts, byte volumes, time-to-live values, jitter metrics, service types, and protocol states.

| | No. | Name | Type | Description |
|---|---|---|---|---|
| 0 | 1 | srcip | nominal | Source IP address |
| 1 | 2 | sport | integer | Source port number |
| 2 | 3 | dstip | nominal | Destination IP address |
| 3 | 4 | dsport | integer | Destination port number |
| 4 | 5 | proto | nominal | Transaction protocol |
| 5 | 6 | state | nominal | Indicates to the state and its dependent proto… |
| 6 | 7 | dur | Float | Record total duration |
| 7 | 8 | sbytes | Integer | Source to destination transaction bytes |
| 8 | 9 | dbytes | Integer | Destination to source transaction bytes |
| 9 | 10 | sttl | Integer | Source to destination time to live value |
| 10 | 11 | dttl | Integer | Destination to source time to live value |
| 11 | 12 | sloss | Integer | Source packets retransmitted or dropped |
| 12 | 13 | dloss | Integer | Destination packets retransmitted or dropped |
| 13 | 14 | service | nominal | http, ftp, smtp, ssh, dns, ftp-data ,irc and … |
| 14 | 15 | Sload | Float | Source bits per second |
| 15 | 16 | Dload | Float | Destination bits per second |
| 16 | 17 | Spkts | integer | Source to destination packet count |
| 17 | 18 | Dpkts | integer | Destination to source packet count |
| 18 | 19 | swin | integer | Source TCP window advertisement value |
| 19 | 20 | dwin | integer | Destination TCP window advertisement value |
| 20 | 21 | stcpb | integer | Source TCP base sequence number |
| 21 | 22 | dtcpb | integer | Destination TCP base sequence number |
| 22 | 23 | smeansz | integer | Mean of the ?ow packet size transmitted by the… |
| 23 | 24 | dmeansz | integer | Mean of the ?ow packet size transmitted by the… |
| 24 | 25 | trans_depth | integer | Represents the pipelined depth into the connec… |
| 25 | 26 | res_bdy_len | integer | Actual uncompressed content size of the data t… |
| 26 | 27 | Sjit | Float | Source jitter (mSec) |

Fig.2 Dataset

The target variable is a binary class label, where each record is classified as either normal or malicious. Additionally, an attack category attribute is provided, identifying various attack types such as fuzzers, exploits, reconnaissance, and denial-of-service attacks. A significant class imbalance is observed, with normal traffic representing most instances—an intentional design choice that mirrors real-world network conditions.

Prior to modeling, duplicate records are removed to reduce bias, and missing values are handled based on domain interpretation. Categorical features are standardized to correct encoding inconsistencies, while numerical features are inspected for skewness and outliers. These preprocessing steps ensure data consistency and reliability, resulting in a structured dataset suitable for supervised machine learning–based threat detection.

## 3. Methodology

The suggested threat detection model adheres to a sequence of steps, which are data preprocessing, feature preparation, neural network modeling, and performance evaluation. This aim is to properly categorize the network traffic flows as benign or malicious in a social media network environment.

Preparation of data and features Preference

**4.1 Data Preprocessing Preference Feature:** Preparation Preference Elimination of repeated records to avoid redundancy and overfitting opens the preprocessing stage. Missing values are handled by domain specific assumptions; missing indicators in protocol-based features are handled by assuming the lack of the relevant network behavior in place of data corruption. Categorical variables are purged, standardized, and coded into numerical value forms that can be used to train neural networks [17].

To counter the effects of extreme values, statistical methods are used to manage the outliers of the individual numerical characteristics chosen. Normalization procedures are applied to feature scaling to provide uniform contribution of all features when training the model. Since the dataset has a strong class imbalance, resampling techniques will be used to enhance the capability of the model to identify minority classes of attack without affecting the performance of the overall model.

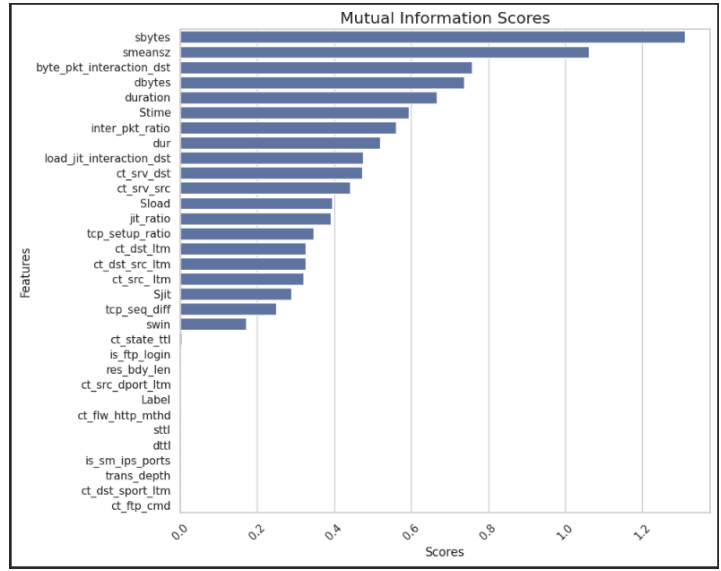

Fig.3 MI Score

**4.2 Neural Network Architecture:** The type of classification model used is the Artificial Neural Network (ANN) because it can provide complex, non-linear association between high-dimensional network traffic characteristics. The network is built up of several fully connected layers that are meant to learn more high-level abstractions of the input data.

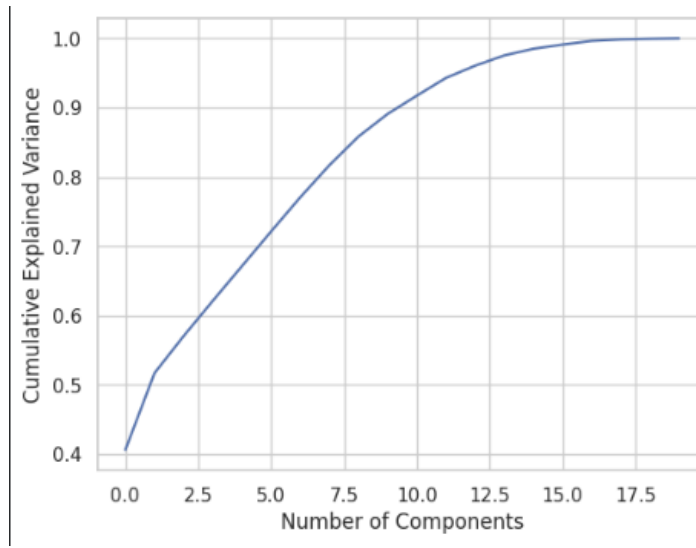

Fig.4 Cumulative Explained Variance

Between the layers, non-linear activation functions are used to increase representational capacity and regularization mechanisms like dropout and batch normalization are added to reduce overfitting and increase generalization. The supervised learning method is used to train the model with a proper loss function optimized through gradient-based optimization algorithms. Empirical choices of hyperparameters are made based on the stability of the performance: such parameters as learning rate, number of hidden layers, and the number of neurons[6]. The training and evaluation strategy will be developed in alignment with the training and development strategy.

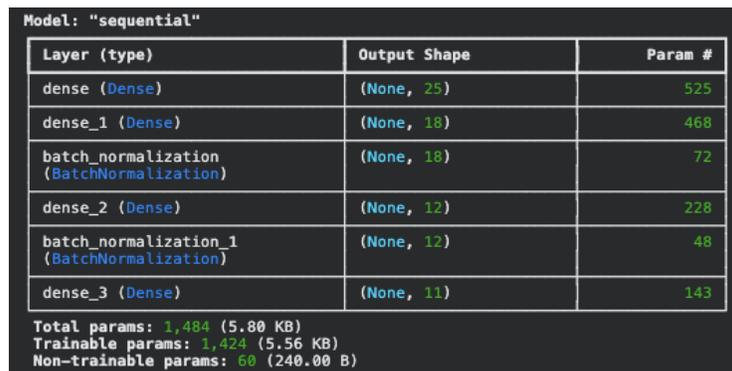

Fig. CNN Layer

**4.3 Training and Evaluation Strategy**: This will be prepared in line with the training and development strategy. The dataset is separated into training and testing data sets to test the capability of generalization. Performance of the models is measured as based on commonly used classification measures, such as accuracy, precision, recall, F1-score, and ROC–AUC. These measures enable a detailed analysis of the detection efficacy especially when the classification is imbalanced as is the case in cybersecurity applications [14,15]. Scalability, robustness, and adaptability are the key features of the proposed methodology that makes it applicable to deployment in the large-scale social media network environment where the rapid and accurate threat detection should be paramount [16].

## 5 . Results and Discussion

In this part, the experimental findings of the suggested ANN-based threat detection model are provided and its efficiency in detecting malicious network traffic in the context of social networks discussed [3,4]. To make sure that the model performance is robust and interpretable, standard classification metrics and visual diagnostic tools are used to assess the model's performance.

**5.1 General Model Performance Evaluation**: It is seen that the trained artificial neural network has good performance in terms of discriminating between normal and malicious network traffic. The evaluation is performed on the held-out test set with a combination of multiple metrics to consider the class imbalance and the reliability of detection. the proposed model when compared with respect to accuracy, precision, recall, F1-score, and ROC/AUC.

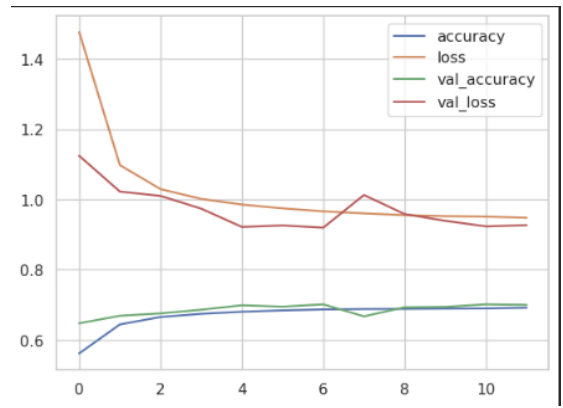

Fig.5 Accuracy, Loss Graph

The value of the recall is high, pointing at the fact that the model is quite successful in recognizing the threats of malicious traffic, which is especially important in the use of the model in cybersecurity, where the threat of unidentified attacks can be hazardous. Precision is also competitive and indicates that the false positives are quite controlled even though the dataset is asymmetrical. The fact that the F1-score indicates a trade-off between detection sensitivity and prediction reliability is further supported by the fact that the F1-score was a balance in both. The ROC-AUC score indicates the high discriminative power of the model on different decision threshold levels indicating the model is robust enough to draw a line between normal and attack traffic patterns.

**Confusion Matrix Analysis:**

To gain deeper insight into classification behavior, a confusion matrix is analyzed, as shown in Figure 6

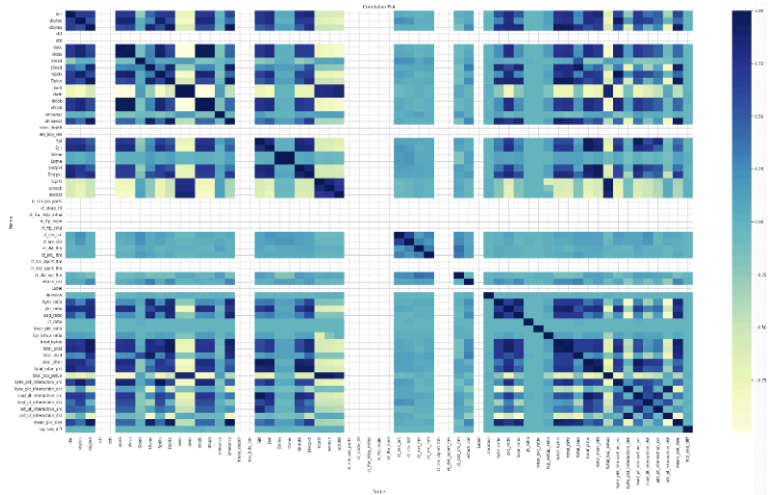

Fig.6 Confusion Matrix

The confusion matrix shows that most malicious traffic cases are correctly identified, and the number of false negatives is not so high. This is what intrusion detection systems would want to achieve because the lower false negativity, the less chances of attacks being unnoticed. Although false positives are expected, such kind of behavior is tolerable in security sensitive environments where cautious flagging is a common characteristic than missed threats[11-13].

**5.4 Discussion and Interpretation:**
There are few points of observation as brought out in the experimental results. To begin with, the ANN is highly efficient in complex and non-linear capturing of relationships among features of network traffic, and thus, it can detect threats in a highly imbalanced dataset. This confirms the previous evidence that the use of deep learning models is better in capturing the complex network behavior compared to traditional machine learning methodologies. Second, unlike content-based threat detection methods that are typically applied in the context of analyzing social media, which include sentiment or text-based detection, the suggested framework can work on the network traffic level and is robust to manipulation with language, semantic ambiguity, and adversarial obfuscation of content. The latter attribute renders the model especially appropriate as an auxiliary protective measure in social media networks. Lastly, due to its scalability, which is coupled with powerful preprocessing and feature management, the framework may be applicable in actual social media network settings where high-traffic rates and changing attack patterns are frequent.

**Conclusion and Future Work**

This paper offered a machine learning-powered threat detection model tailored to social media networks settings through the artificial neural network (ANN) model. Using fined-tuned network traffic features and extensive preprocessing chain, the suggested method can efficiently detect malicious traffic in large and unbalanced network data. Data consistency and strength was ensured through extensive exploration analysis and feature preparation before the training of models.

The experimental outcomes reveal that ANN model has high level of detection performance in various evaluation measures such as accuracy, precision, recall, F1-score and ROC-AUC. The model also has a very high recall rate, which means that it is effective at detecting malicious traffic and reducing the number of threats that go unnoticed, and that is a critical need of cybersecurity application. The reliability of the model as well as its discriminative ability is also validated by the confusion matrix analysis and ROC curve[5].

In contrast to content-based threat detection mechanisms, which assume textual or sentiment analysis, the outlined framework is network traffic based so that it is not susceptible to language diversity, manipulations in the semantics, and content obfuscation that tend to be present on social media platforms. This additional viewpoint supplements the overall security stance of social media systems because it allows us to identify threats in advance and on a behavioral basis.

Even though the findings are encouraging, there are several ways in which it can be taken forward. The present research is on binary classification; it may be expanded to multi-class attack classification to get more specific threat information. This can be enhanced by adding ensemble learning methods or combined deep learning frameworks that could enhance detection accuracy and resiliency[7,9,10]. Moreover, the explainable AI practices may also contribute to a higher level of interpretability of the model, which will make it easier to embrace and trust operational security settings. Future studies can also be conducted, examining the actual deployment scenarios as well as testing the performance of the framework under the conditions of live social media traffic.

Comprehensively, this paper adds to the existing body of research in intelligent intrusion detection systems, and it shows that neural network-based models are applicable in detecting the threat scalable in the current social media network setting.